\documentclass[11pt]{article}

\usepackage[preprint]{acl}

\usepackage{times}
\usepackage{latexsym}

\usepackage[T1]{fontenc}

\usepackage[utf8]{inputenc}

\usepackage{microtype}

\usepackage{inconsolata}

\usepackage{graphicx}

\usepackage{amsmath}
\usepackage{amsfonts}
\usepackage{multirow}
\usepackage{subfigure}
\usepackage{xspace}
\usepackage{enumitem} 
\usepackage[hypcap=true]{caption}
\usepackage{comment}
\usepackage{booktabs}
\usepackage{amsthm}
\usepackage{colortbl}
\usepackage[table]{xcolor}

\newcommand{\mname}{{\sc POET}\xspace}
\newtheorem{theorem}{Theorem}

\definecolor{basebg}{HTML}{F5F5F5}     
\definecolor{methodbg}{HTML}{E3F2FD}    
\definecolor{bestcolor}{HTML}{1565C0} 
\newcommand{\best}[1]{\textbf{#1}}

\definecolor{heat1}{HTML}{F7FBFF}  
\definecolor{heat2}{HTML}{DEEBF7}  
\definecolor{heat3}{HTML}{C6DBEF}  
\definecolor{heat4}{HTML}{9ECAE1}  
\definecolor{heat5}{HTML}{6BAED6}  
\definecolor{heat6}{HTML}{4292C6}  

\newcommand{\heatA}[1]{\cellcolor{heat1}#1}
\newcommand{\heatB}[1]{\cellcolor{heat2}#1}
\newcommand{\heatC}[1]{\cellcolor{heat3}#1}
\newcommand{\heatD}[1]{\cellcolor{heat4}#1}
\newcommand{\heatE}[1]{\cellcolor{heat5}#1}
\newcommand{\heatF}[1]{\cellcolor{heat6}#1}

\title{Towards Bridging the Reward-Generation Gap \\ in Direct Alignment Algorithms}

\author{
    Zeguan Xiao$^1$,
    Yun Chen$^{1}$,
    Jian Yang$^2$,
    Guanhua Chen$^3$\thanks{\ \ Corresponding authors.},
    Ke Tang$^3$\footnotemark[1] \\
    $^1$Shanghai University of Finance and Economics, $^2$Beihang University\\
    $^3$Southern University of Science and Technology
}

\begin{document}
\maketitle
\begin{abstract}
Direct Alignment Algorithms (DAAs), such as Direct Preference Optimization (DPO) and Simple Preference Optimization (SimPO), have emerged as efficient alternatives to Reinforcement Learning from Human Feedback (RLHF) algorithms for aligning large language models (LLMs) with human preferences.
However, DAAs suffer from a fundamental limitation we identify as the ``reward-generation gap''—a discrepancy between training objectives and autoregressive decoding dynamics. In this paper, we consider that one contributor to the reward-generation gap is the mismatch between the inherent importance of prefix tokens during the LLM generation process and how this importance is reflected in the implicit reward functions of DAAs.
To bridge the gap, we adopt a token-level MDP perspective of DAAs to analyze its limitations and introduce a simple yet effective approach called \textbf{P}refix-\textbf{O}riented \textbf{E}qual-length \textbf{T}raining (POET), which truncates both preferred and dispreferred responses to match the shorter one's length.
We conduct experiments with DPO and SimPO, two representative DAAs, demonstrating that \mname improves over their standard implementations, achieving up to 11.8 points in AlpacaEval 2 and overall improvements across downstream tasks.
These results underscore the need to mitigate the reward-generation gap in DAAs by better aligning training objectives with autoregressive decoding dynamics.
\end{abstract}

\section{Introduction}

\begin{figure}[htbp]
    \centering
    \includegraphics[width=0.98\columnwidth]{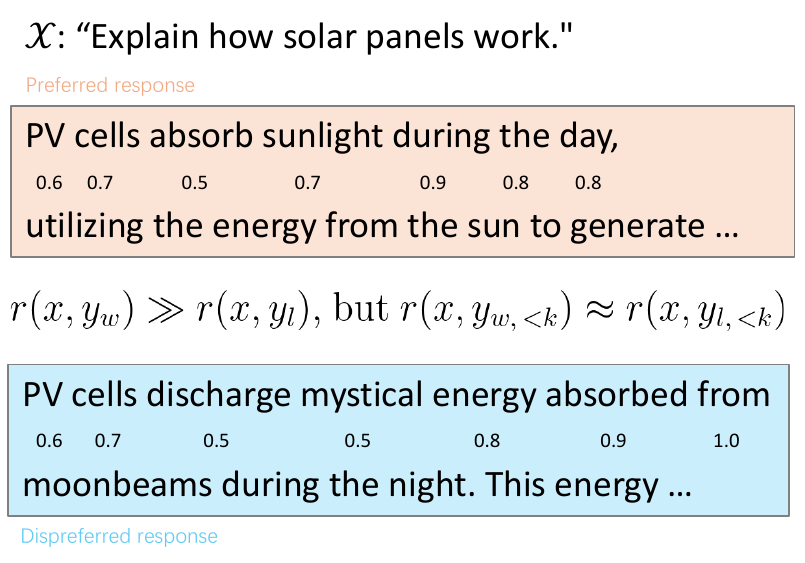}
    \caption{During autoregressive generation, LLMs generate tokens sequentially from left to right. Although DAAs are optimized to ensure $r(x, y_w) \gg r(x, y_l)$ over entire sequences, they do not guarantee that $r(x, y_{w,<k}) \gg r(x, y_{l,<k})$ for prefixes.
    }
    \label{fig:intro_issue}
\end{figure}

Large language models (LLMs)~\cite{Brown2020LanguageMA,touvron2023llama,llama3modelcard} have demonstrated remarkable capabilities across a wide range of tasks, including instruction following~\cite{zhou2023instruction}, mathematical problem solving~\cite{cobbe2021gsm8k,shao2024deepseekmath}, and coding generation~\cite{chen2021evaluating,roziere2023code}.
An important step in the training of LLMs is ``alignment'', which refers to aligning LLMs with human intentions and values, ensuring they are helpful, honest, and harmless~\citep{Askell2021AGL}. 
Learning from human feedback plays a crucial role in LLM alignment, and a popular paradigm for this is Reinforcement Learning with Human Feedback (RLHF)~\citep{christiano2017deep, ziegler2019fine, Ouyang2022TrainingLM, bai2022training}. The RLHF pipeline typically involves a three-stage process: supervised fine-tuning, reward modeling, and policy optimization. While effective, RLHF faces challenges including training instability, computational inefficiency, and high sensitivity to hyperparameters~\citep{zheng2023secrets, santacroce2023efficient}, motivating the exploration of alternative approaches.

To simplify the alignment process, researchers have developed Direct Alignment Algorithms (DAAs)~\citep{rafailov2024scaling}, such as Direct Preference Optimization (DPO)~\citep{Rafailov2023DirectPO} and Simple Preference Optimization (SimPO)~\citep{meng2024simpo}. DAAs bypass the need for explicit reward modeling and reinforcement learning in RLHF, instead directly performing preference optimization based on a reference dataset. The simplicity and effectiveness of DAAs have made them an attractive alternative to RLHF.

The DAAs are optimized to increase the reward of preferred responses while reducing the reward of dispreferred ones~\citep{Ethayarajh2024KTOMA,Azar2023AGT,Zhao2023SLiCHFSL}, where some can be expressed as implicit reward functions based on different manifestations of likelihoods~\citep{Rafailov2023DirectPO,meng2024simpo}.
However, a known issue with DAAs is that they may decrease the reward of preferred responses as long as the reward margins between preferred and dispreferred responses increase~\citep{Rafailov2023DirectPO,pal2024smaug}. More importantly, \citet{shi2024understanding} found that neither a higher reward of preferred responses nor larger reward margins necessarily lead to better performance, suggesting a deeper issue in how DAAs connect optimization objectives to LLM's performance.

In this paper, we identify this issue as the \textbf{Reward-Generation Gap in DAAs}—a fundamental misalignment between DAAs' training objectives and the autoregressive decoding dynamics of LLMs.
To bridge this gap, we adopt a token-level MDP perspective of DAAs to analyze their limitations.
Our subsequent theoretical and empirical analysis provides the motivation for our proposed method.
Building on these insights, we propose a simple yet effective data augmentation approach, \textbf{P}refix-\textbf{O}riented \textbf{E}qual-length \textbf{T}raining (\mname), which truncates both preferred and dispreferred responses in each sample to match the shorter one's length, resulting in diverse truncated lengths across samples. Training with \mname, the optimization of DAAs' objective is implicitly constrained to converge across all timesteps of token-level MDP.

We conduct extensive experiments with DPO and SimPO under various experimental settings, demonstrating that \mname consistently improves over their standard implementations. \mname achieves up to 11.8 points gain on AlpacaEval 2~\citep{AlpacaEval,dubois2024length}. Experiments on downstream tasks also show overall improvements across various benchmarks.
Our analysis further reveals that \mname effectively bridges the reward-generation gap by generating better prefixes.\footnote{Our code is publicly available at \url{https://github.com/sustech-nlp/POET}.}

\section{Reward-Generation Gap in DAAs}
\label{sec:reward-generation-gap}

\subsection{Background} \label{sec:background}

In RLHF~\citep{Ouyang2022TrainingLM,stiennon2020learning,li-etal-2016-deep,ziegler2019fine}, LLM learns a policy $\pi_\theta$, and $\pi_\theta(y \mid x)$ represents the probability assigned by the LLM to a response $y$ given an input prompt $x$.
During the reinforcement learning (RL) phase, the optimization objective is to maximize the expected reward while preventing the policy $\pi_\theta$ from deviating too far from the reference policy $\pi_{\mathrm{ref}}$:
\begin{equation}
\label{eq:rl_objective}
\begin{aligned}
\max _{\pi_\theta} \; & \mathbb{E}_{x \sim \mathcal{D},\, y \sim \pi_\theta(y \mid x)}
\left[ r_\phi(x, y) \right] \\
& - \beta\, \mathbb{D}_{\mathrm{KL}}\left[ \pi_\theta(y \mid x) \| \pi_{\mathrm{ref}}(y \mid x) \right],
\end{aligned}
\end{equation}
where $r_\phi$ is the reward model, typically trained as a Bradley-Terry (BT) model using a static preference dataset $\mathcal{D}$ of triples $(x, y_w, y_l)$. In each triple, $x$ denotes the input prompt, $y_w$ the preferred response, and $y_l$ the dispreferred response. The reward model receives the generated response and provides a reward signal $r_\phi(x, y)$ indicating the quality of the response.

Alternatively, DAAs~\citep{rafailov2024scaling} eliminate the need for explicit reward modeling and reinforcement learning, directly performing preference optimization on a static preference dataset.
The optimization objectives of DAAs contain rewards implicitly defined by $\pi_\theta$, where some rewards are based directly on the likelihood~\citep{Zhao2023SLiCHFSL,xu2024contrastive}, while others are derived from different manifestations of likelihoods, such as DPO~\citep{Rafailov2023DirectPO} and SimPO~\citep{meng2024simpo}:
\begin{equation}
\label{eq:dpo_reward}
r_{\text{DPO}}(x,y) = \beta \log \frac{\pi_\theta(y \mid x)}{\pi_{\text{ref}}(y \mid x)},
\end{equation}
\begin{equation}
\label{eq:simpo_reward}
r_{\text{SimPO}}(x, y) = \frac{\beta}{|y|} \log \pi_\theta(y \mid x),
\end{equation}
where $r_{\text{DPO}}(x,y)$ and $r_{\text{SimPO}}(x,y)$ are the implicit reward functions of DPO and SimPO, respectively.

\subsection{Reward-Generation Gap in DAAs}
\label{sec:issue}

Designed to increase the reward of preferred responses $y_w$ while decreasing the reward of dispreferred responses $y_l$, DAAs converge if the reward margins between preferred and dispreferred responses grow sufficiently large. However, a known issue with DAAs is that they may decrease the reward of preferred responses as long as the margins increase~\citep{Rafailov2023DirectPO,pal2024smaug}.
Furthermore, \citet{shi2024understanding} finds that contrary to expectations, neither a higher reward of preferred responses nor larger reward margins between the preferred and dispreferred responses necessarily lead to better performance.

The disconnect identified above arises from the fact that the implicit rewards of DAAs do not represent the quality of responses, but are rather different manifestations of their likelihoods, such as Eq.~\ref{eq:dpo_reward} and Eq.~\ref{eq:simpo_reward}.
However, due to the foundational discrepancy between the maximum/minimum likelihood training and autoregressive generation in language models~\citep{ranzato2015sequence,zhang-etal-2021-trading,arora-etal-2022-exposure}, optimizing for the DAAs training objectives does not necessarily lead to better generation quality.
This discrepancy creates a substantial misalignment between training objectives and autoregressive decoding dynamics—a phenomenon we refer to as the \textbf{Reward-Generation Gap in DAAs}.

We consider that one contributor to this gap is the mismatch between the inherent importance of prefix tokens during the LLM generation process and how this importance is reflected in the implicit reward functions of DAAs.
As shown in Fig.~\ref{fig:token_entropy} in Appendix~\ref{app:token_dynamics}, the token-level entropies (i.e., uncertainty) are significantly higher in the early positions, and gradually decrease as more context becomes available.
As demonstrated by~\citet{arora-etal-2022-exposure}, errors in autoregressive generation tend to accumulate, with early mistakes propagating and amplifying through the sequence. This phenomenon, known as exposure bias, means that suboptimal choices in prefix tokens can severely degrade the quality of the entire response.
However, DAAs' reward functions assign equal weight to every token, ignoring these positional differences.
As illustrated in Fig.~\ref{fig:token_lopg}, the log probability of early tokens is significantly lower than that of later tokens, yet their contribution to DAAs' implicit reward is diluted by the overwhelming number of subsequent tokens.
This is highlighted by the large gap between the early tokens' log probabilities and the mean log probability across all positions, which is a key component of the implicit reward.

\section{How to Bridge the Gap?}
\label{sec:method_and_results}

\subsection{Token-level MDP perspective of DAAs}

Most classical RLHF approaches formulate the preference optimization problem as a contextual bandit problem, where the dataset $\mathcal{D}=\left\{\left(\mathbf{x}^{(i)}, \mathbf{y}^{(i)}\right)\right\}_{i=1}^N$ of language prompts $\mathbf{x}$ and target answers $\mathbf{y}$, each of which can be broken down into a sequence of tokens, for example $\mathbf{x}=\left(x_0, \ldots, x_m\right)$, from a fixed discrete vocabulary $\mathcal{A}$.
In this formulation, the entire response $\mathbf{y}$ is treated as a single action, and the reward is computed based on the entire response. DAAs, such as DPO and SimPO, stay entirely within the contextual bandits setting, but can theoretically be cast into the token-level MDP~\citep{rafailov2024r}.

In the perspective of token-level MDP of DAAs, the state $\mathbf{s}$ consists of all tokens generated so far (i.e., $\mathbf{s}_t = \left( x_0, \ldots, x_m, y_0, \ldots, y_t \right)$), and the action $a$ is to select a single token $y_{t+1}$ from the vocabulary $\mathcal{A}$.
The corresponding Bradley-Terry (BT) model is therefore defined as follows:
\begin{equation}
\begin{aligned}
p^*\left(y_w \succeq y_l\right) = \frac{A_w}{A_w + A_l}
\end{aligned}
\label{eq:bt_model}
\end{equation}
where $A_w = \exp \left(\sum_{i=1}^N r\left(\mathbf{s}_i^w, \mathbf{a}_i^w\right)\right)$ and $A_l = \exp \left(\sum_{i=1}^M r\left(\mathbf{s}_i^l, \mathbf{a}_i^l\right)\right)$.
$r\left(\mathbf{s}_i^w, \mathbf{a}_i^w\right)$ and $r\left(\mathbf{s}_i^l, \mathbf{a}_i^l\right)$ are the token-level rewards for each action in the preferred and dispreferred trajectory, respectively, and $p^*\left(y_w \succeq y_l\right)$ gives the probability that the preferred trajectory $y_w$ is better than the dispreferred trajectory $y_l$.

Although token-level DPO theoretically enables the derivation of an optimal policy $\pi^*$ for the underlying MDP of Eq.~\ref{eq:rl_objective}~\citep{rafailov2024r}\footnote{Other DAAs can also be formulated to yield an optimal policy for an MDP; however, the underlying MDP for them is distinct from that defined by Eq.~\ref{eq:rl_objective}.}, it is challenging to train $\pi^*$ in practice, as we only possess sparse reward signals in the form of sequence-level labels indicating that $y_w$ is preferred over $y_l$.
Furthermore, the sequence-level BT model does not model the preference of sub-trajectories, failing to capture the convergence behavior of partial sequences, which leads to the issue illustrated in Figure~\ref{fig:intro_issue}.

\subsection{Theoretical Basis of Proposed \mname}

Inspired by the limitations of DAAs from the token-level MDP perspective, we introduce a formal theoretical basis for our approach. Our key insight is that optimizing policies on equal-length sub-trajectories yields the same optimal policy as sequence-level optimization, while providing better supervision for crucial early tokens.

We begin by defining the equal-length sub-trajectories BT model:
\begin{equation}
\label{eq:el_bt_model}
p^*_k\left(y_{w,\leq k} \succeq y_{l,\leq k}\right) = \frac{E_w}{E_w + E_l}
\end{equation}
where $E_w = \exp \left(\sum_{t=0}^{k} r\left(\mathbf{s}_t^w, \mathbf{a}_t^w\right) + V^*(\mathbf{s}_{k+1}^w)\right)$ and $E_l = \exp \left(\sum_{t=0}^{k} r\left(\mathbf{s}_t^l, \mathbf{a}_t^l\right) + V^*(\mathbf{s}_{k+1}^l)\right)$.
$y_{w,\leq k} \succeq y_{l,\leq k}$ indicates that the sub-trajectory $y_{w,\leq k}$ is preferred over $y_{l,\leq k}$ when the cumulative reward plus the state value of the resulting state for $y_{w,\leq k}$ exceeds that of $y_{l,\leq k}$. Here, $V^*$ represents the optimal state-value function under the MDP defined by Eq.~\ref{eq:rl_objective}.

\begin{theorem}
\label{thm:optimality}
The policy derived from the optimal equal-length sub-trajectory BT model is equivalent to the optimal policy derived from the original sequence-level BT model defined in Eq.~\ref{eq:bt_model} for DPO.
\end{theorem}

The proof strategy is to show that the optimal equal-length sub-trajectories BT model can be expressed in terms of the optimal policy from the original sequence-level optimization, demonstrating that the two approaches yield equivalent optimal policies.
The detailed proof is provided in Appendix~\ref{appendix:theorem_proof}\footnote{The proof for SimPO follows a similar structure, but with a different expression of the BT model using the optimal policy.}.

\subsection{Feasibility of Equal-Length Preference Training}
\label{sec:prefix_importance}

While Theorem~\ref{thm:optimality} establishes that the optimal policy is preserved unconditionally when optimizing the equal-length sub-trajectory BT model, in practice we face two challenges: (1) existing preference datasets only provide full-sequence preference labels indicating that $y_w \succ y_l$, rather than off-the-shelf equal-length preference data; (2) we cannot directly compute the optimal state-value function $V^*$ required by the BT model. In this section, we empirically investigate the feasibility of directly reusing existing full-sequence preference data for training.

We begin with a central observation: \textbf{the quality of a response to be generated heavily depends on the quality of its initial prefix}.
Formally, we denote a prefix of length $k$ as $y_{\leq k}$ and define the quality of $y_{\leq k}$ as the expected quality of complete responses given it:
\begin{equation}
\label{eq:prefix_quality}
Q(y_{\leq k}) = \mathbb{E}_{y_{\textgreater k} \sim \pi_\theta(y_{\textgreater k} \mid x, y_{\leq k})} \left[ r_\theta(x, y_{\leq k} \oplus y_{\textgreater k}) \right],
\end{equation}
where $\pi_\theta$ is an oracle policy, $r_\theta$ is the oracle reward model that evaluates the quality of responses, $y_{\textgreater k}$ denotes the completion given prefix $y_{\leq k}$, and $\oplus$ represents the concatenation operation.

\paragraph{Experimental setup.}
We randomly select 1,000 samples from the training set of UltraFeedback~\citep{Cui2024UltraFeedbackBL}.
For each pair of responses $(y_w, y_l)$ in these samples, we truncate both responses at different positions $k$ to obtain prefixes of varying lengths, and then generate multiple completions from these prefixes. We use two models as proxy policy $\hat{\pi}_\theta$: Zephyr~\citep{Tunstall2023ZephyrDD} and Llama-3-Base-8B-SFT~\citep{meng2024simpo}.
We quantify the prefix quality difference by $\Delta Q(k) = Q(y_{w,\,\leq k}) - Q(y_{l,\,\leq k})$ at different prefix lengths $k$, utilizing ArmoRM-Llama3-8B-v0.1~\citep{ArmoRM} as proxy reward model.

\begin{figure}[t]
    \centering
    \includegraphics[width=0.48\textwidth]{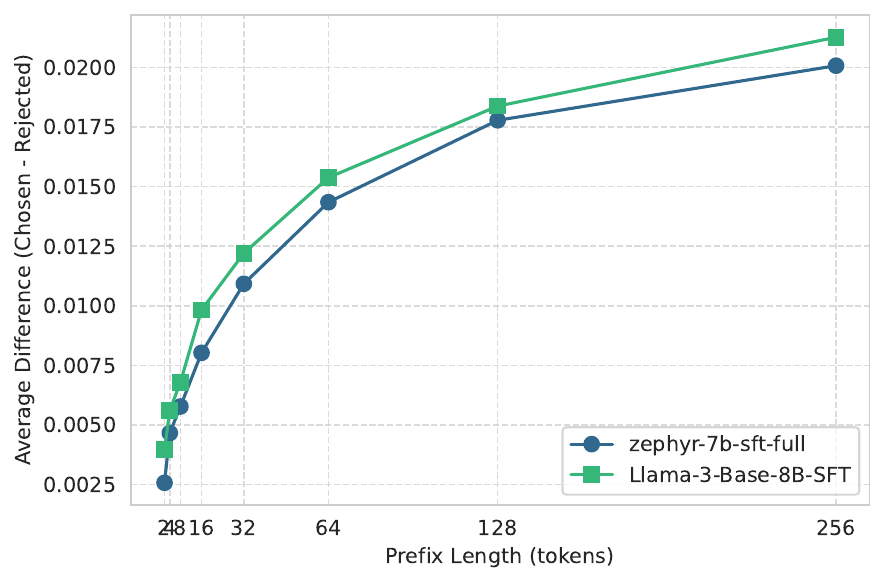}
    \caption{Average prefix quality difference between preferred and dispreferred responses at different prefix lengths. The results demonstrate that quality differences emerge early in the prefixes and increase with prefix length, though with diminishing marginal gains at longer lengths.}
    \label{fig:prefix_rewards}
\end{figure}

\paragraph{Results.}
As shown in Figure~\ref{fig:prefix_rewards}, the prefix quality gap emerges very early and grows as the prefix length increases, with diminishing marginal gains at longer lengths. Specifically, the marginal increase $\Delta Q(k+1) - \Delta Q(k)$ decreases substantially as $k$ increases and becomes negligible for sufficiently large $k$, indicating that $\Delta Q(k)$ approaches a plateau. This convergence behavior implies that for sufficiently large $k$, the difference $V^*(\mathbf{s}_{k+1}^w) - V^*(\mathbf{s}_{k+1}^l)$ becomes negligible, confirming that the full-sequence preference ordering is preserved after equal-length truncation.

\textbf{From the token-level MDP perspective, the majority of reward is obtained in the early timesteps, after which the trajectory enters a stationary phase and the incremental reward difference from subsequent actions diminishes significantly}.

\subsection{Prefix-Oriented Equal-length Training}
\label{sec:method}

We introduce \mname, a straightforward and effective method designed to prompt DAAs to capture the convergence behavior of partial sequences, thereby bridging the reward-generation gap.

Motivated by these theoretical and empirical findings, we propose \mname to approximate the optimization of the equal-length sub-trajectories BT model.
Concretely, given a pair of preferred and dispreferred responses $(y_w, y_l)$ with lengths $|y_w|$ and $|y_l|$, we truncate both responses to the length of the shorter one, denoted as $k = \min(|y_w|, |y_l|)$, resulting in truncated responses $y_{w,\,\leq k}$ and $y_{l,\,\leq k}$.
By Theorem~\ref{thm:optimality}, optimizing the equal-length sub-trajectory BT model yields the same optimal policy as full-sequence optimization. In practice, since $k = \min(|y_w|, |y_l|)$, one response remains complete while only the suffix of the longer one is discarded. As shown empirically in Section~\ref{sec:prefix_importance}, the quality ranking is preserved after truncation with high consistency (Table~\ref{tab:truncate_acc}), validating the use of full-sequence preference labels for the truncated equal-length pairs.

Training with \mname, where both responses in each sample are equal in length and diverse lengths across samples, the optimization of DAAs' objective is implicitly constrained to converge across timesteps of token-level MDP, thus providing finer-grained reward signals of sub-trajectories and paying more attention to prefix tokens than the standard DAAs.

We summarize three advantages of \mname:
\begin{itemize}[leftmargin=*]
    \item \textbf{Universal compatibility:} \mname is compatible with any DAAs, requiring no modifications to their optimization objectives. This allows for seamless integration with both current and future DAAs, making \mname a versatile enhancement to DAAs.
    \item \textbf{Hyperparameter-free:} By using the shorter response's length as a natural truncation point, \mname requires no additional hyperparameters. This makes implementation straightforward and eliminates the need for additional hyperparameter tuning.
    \item \textbf{Minimal data noise risk:} There is a possibility that after truncation, the originally preferred response might become inferior to the dispreferred one, i.e., $Q(y_{w,\,\leq k}) < Q(y_{l,\,\leq k})$. \mname inherently minimizes the risk of introducing noise into training data. By truncating to the shorter response's length, we ensure that the risk only comes from the truncated suffix of the longer response, which is less critical to the overall response quality according to our empirical analysis in Figure~\ref{fig:prefix_rewards}. We verify the risk empirically in three settings and report the results in Table~\ref{tab:truncate_acc}. Specifically, we get $y_{w,\,\leq k}$ and $y_{l,\,\leq k}$ by \mname, generate completions from the truncated prefix, and evaluate whether the quality ranking between the completion and the untruncated shorter response is preserved, using ArmoRM-Llama3-8B-v0.1~\citep{ArmoRM} as the reward model. As shown in Table~\ref{tab:truncate_acc}, the quality consistency before and after truncation is high across all three settings: 98.5\% in the Llama-3-8B-Instruct v0.2 (on-policy)~\citep{meng2024simpo} setting, 93.8\% in the v0.1 (on-policy) setting (lower due to a weaker reward model for annotation), and 91.4\% in the Zephyr + UltraFeedback (off-policy) setting, where Zephyr's behavior differs from the data-generating model, making it a less ideal proxy policy.
\end{itemize}

\begin{table}[t]
    \centering
    \begin{tabular}{lc}
      \toprule
      Setting & Consistency \\
      \midrule
      Zephyr + UltraFeedback & 91.4\% \\
      Llama-3-8B-Instruct v0.1 & 93.8\% \\
      Llama-3-8B-Instruct v0.2 & 98.5\% \\
      \bottomrule
    \end{tabular}
    \caption{Quality ranking consistency before and after truncation across three settings.}
    \label{tab:truncate_acc}
\end{table}

\section{Experiments}

\subsection{Experimental Setup}
\label{sec:exp_setup}

\begin{table*}[t]
    \centering
    \resizebox{\textwidth}{!}{
    \begin{tabular}{lcccccccc}
    \toprule
    \multirow{3}{*}{\textbf{Method}} & \multicolumn{4}{c}{\textbf{Mistral-Base (7B)}} & \multicolumn{4}{c}{\textbf{Llama-3-Base (8B)}} \\
    \cmidrule(lr){2-5}\cmidrule(lr){6-9}
    & \multicolumn{3}{c}{\bf AlpacaEval 2} & {\bf Arena-Hard} & \multicolumn{3}{c}{\bf AlpacaEval 2} & {\bf Arena-Hard} \\  
    \cmidrule(lr){2-4}\cmidrule(lr){5-5}\cmidrule(lr){6-8}\cmidrule(lr){9-9}
    & {\bf LC (\%)} & {\bf WR (\%)} & {\bf Length} & {\bf WR (\%)} & {\bf LC (\%)} & {\bf WR (\%)} & {\bf Length} & {\bf WR (\%)} \\
    \midrule
    \rowcolor{basebg} SFT & 5.3 & 5.3 & 931 & 2.6 & 6.1 & 4.0 & 976 & 6.0 \\
    \midrule
    DPO & 12.9 & 10.6 & 1569 & 11.2 & 16.9 & 14.4 & 1644 & 18.6 \\
    \rowcolor{methodbg} + \mname & \best{24.7} & \best{17.7} & 1401 & \best{14.6} &  \best{28.4} & \best{21.4} & 1403 & \best{25.5} \\
    \midrule
    SimPO & 20.0 & 18.0 & 1777 & 16.0 & 28.0 & 25.3 & 1777 & 13.5 \\
    \rowcolor{methodbg} + \mname & \best{24.2} & \best{23.7} & 1939 & \best{17.4} & \best{33.8} & \best{32.3} & 1964 & \best{14.8} \\
    \midrule[.7pt]
    \multirow{3}{*}{\textbf{Method}} & \multicolumn{4}{c}{\textbf{Llama-3-Instruct v0.2 (8B)}} & \multicolumn{4}{c}{\textbf{Gemma-2-Instruct (9B)}} \\
    \cmidrule(lr){2-5}\cmidrule(lr){6-9}
    & \multicolumn{3}{c}{\bf AlpacaEval 2} & {\bf Arena-Hard} & \multicolumn{3}{c}{\bf AlpacaEval 2} & {\bf Arena-Hard} \\
    \cmidrule(lr){2-4}\cmidrule(lr){5-5}\cmidrule(lr){6-8}\cmidrule(lr){9-9}
    & {\bf LC (\%)} & {\bf WR (\%)} & {\bf Length} & {\bf WR (\%)} & {\bf LC (\%)} & {\bf WR (\%)} & {\bf Length} & {\bf WR (\%)} \\
    \midrule
    \rowcolor{basebg} SFT & 28.4 & 27.4 & 1932 & 19.0 & 52.6 & 39.9 & 1558 & 45.4 \\
    \midrule
    DPO & 65.9 & \best{63.3} & 1998 & \best{36.3} & 78.4 & \best{76.1} & 2026 & 61.2 \\
    \rowcolor{methodbg} + \mname & \best{70.4} & 58.2 & 1615 & 29.1 & \best{79.7} & 72.2 & 1756 & \best{61.7} \\
    \midrule
    SimPO & 68.1 & \best{62.4} & 1805 & \best{29.9} & 78.5 & 73.7 & 1854 & 58.6 \\
    \rowcolor{methodbg} + \mname & \best{70.1} & 57.9 & 1609 & 25.7 & \best{80.1} & \best{74.5} & 1860 & \best{61.6} \\
    \bottomrule
    \end{tabular}
    }
    \caption{Instruction-following evaluation results. LC and WR denote length-controlled and raw win rate, respectively. Length indicates the average length of the responses.}
    \label{tab:main_result}
\end{table*}

\paragraph{Models and training settings.}

Following~\citet{meng2024simpo}, we perform preference optimization under two setups: Base and Instruct.
For the \textbf{Base setup}, we use either Zephyr~\citep{Tunstall2023ZephyrDD} or Llama-3-Base-8B-SFT~\citep{meng2024simpo} as the starting point for preference optimization. These models are trained on the UltraChat-200k dataset~\citep{Ding2023EnhancingCL} and are based on Mistral-7B-v0.1~\citep{jiang2023mistral7b} and Meta-Llama-3-8B~\citep{llama3modelcard}, respectively. The preference optimization is performed on the UltraFeedback dataset~\citep{Cui2024UltraFeedbackBL}, which contains approximately 61K human preference triples.
For the \textbf{Instruct setup}, we use either Meta-Llama-3-8B-Instruct~\citep{llama3modelcard} or gemma-2-9b-it model~\cite{team2024gemma} as the starting point for preference optimization. The responses in the preference datasets for this setup are regenerated by these models using the prompts from the UltraFeedback dataset, bringing the setup closer to an \textit{on-policy} setting. In our experiments, we reuse the preference datasets from~\citet{meng2024simpo}.

As highlighted in~\citet{meng2024simpo}, tuning hyperparameters is critical for achieving optimal performance with all DAAs. Accordingly, in our experiments, we follow the choices of hyperparameters as described in~\citet{meng2024simpo}; further details are provided in Appendix~\ref{app:hyperparameters}.

\paragraph{Baselines.}
We apply \mname to two established DAAs: DPO~\citep{Rafailov2023DirectPO} and SimPO~\citep{meng2024simpo}, which represent reference-based and reference-free DAAs, respectively. Our baselines consist of models trained with these same DAAs, establishing a direct comparison to isolate the impact of \mname.
Additionally, we include SFT models as reference points to assess the relative improvements of all preference optimization methods.

\paragraph{Evaluations.}
We primarily evaluate models along two distinct dimensions of capability: instruction-following ability and performance across diverse downstream tasks.
\textbf{For instruction-following evaluation}, we utilize two widely-adopted benchmarks by the community: AlpacaEval 2~\cite{AlpacaEval} and Arena-Hard v0.1~\cite{arenahard2024}. AlpacaEval 2 provides 805 diverse instructions from 5 datasets, while Arena-Hard v0.1 consists of 500 well-defined technical problem-solving queries.
For AlpacaEval 2, we report both raw win rate (WR) and length-controlled win rate (LC)~\cite{dubois2024length}, with the latter metric designed to control for potential bias from model verbosity. The LC metric has a Spearman correlation of 0.98 with ChatBot Arena~\citep{zheng2023judging}, compared to 0.93 for the WR, making it a more reliable metric for evaluating instruction-following performance. For Arena-Hard, we report the win rate (WR).
\textbf{For assessing downstream task capabilities}, we utilize the comprehensive suite of benchmarks from the Huggingface Open Leaderboard~\citep{open-llm-leaderboard}, including tasks such as GSM8K~\citep{cobbe2021gsm8k}, MMLU~\citep{hendryckstest2021}, and others. We report the average performance across all tasks.
The details of the evaluations can be found in Appendix~\ref{app:downstream_tasks}.
We also conduct experiments on the safety alignment task, observing substantial improvements in safety rates (Section~\ref{sec:safety_alignment}).

\subsection{Main results}
\label{sec:exp_results}

\paragraph{\mname consistently improves the instruction-following performance.}

As shown in Table~\ref{tab:main_result}, \mname consistently improves AlpacaEval 2 LC of both DPO and SimPO without introducing any additional hyperparameters or requiring further hyperparameter tuning. Most notably, \mname achieves a substantial improvement of 11.8 points in AlpacaEval 2 LC for Mistral-Base-7B-DPO. These consistent improvements across all experimental settings demonstrate the robustness and effectiveness of \mname.
In some settings, we observe that while \mname improves the LC, the WR decreases. This contradiction can be attributed to the known bias toward verbosity~\citep{dubois2023alpacafarm,singhal2023long}. As shown in Table~\ref{tab:main_result}, settings where \mname achieves lower WR typically exhibit reduced average response length compared to their baseline counterparts. The WR might favor longer generations due to the absence of a length penalty.
Importantly, in most settings, improvements are maintained across both the LC and WR metrics, suggesting that \mname genuinely enhances response quality rather than merely exploiting metric-specific characteristics.

\paragraph{\mname does not increase the alignment tax on downstream tasks.}
An important consideration for preference optimization methods is the \emph{alignment tax}—the potential degradation of general capabilities as a side effect of alignment training. To verify that \mname does not exacerbate this issue, we evaluate model performance across various downstream tasks (Table~\ref{tab:all_downstream}). Comparing DPO and SimPO models with and without \mname, we observe that \mname does not degrade general capabilities and even yields modest but consistent improvements in overall performance in most settings.
We attribute these moderate gains to the generation of high-quality prefixes in downstream tasks, aligning with findings of~\citet{ji2025first} on the role of prefixes in complex reasoning tasks.
Note that these downstream tasks are not targeted by preference optimization; rather, this evaluation serves to confirm that \mname's gains in instruction-following do not come at the cost of general task performance.

\begin{table}[t]
    \centering
    \resizebox{\columnwidth}{!}{
    \begin{tabular}{llcccc}
    \toprule
    Method & Length Strategy & 25\% & 50\% & 75\% & 100\% \\
    \midrule
    \multirow{2}{*}{DPO} & Original Len. & \heatA{14.1} & \heatB{17.2} & \heatB{16.2} & \heatA{12.9} \\
    & \mname Len. & \heatD{23.5} & \heatE{24.9} & \heatF{26.7} & \heatE{24.7} \\
    \midrule
    \multirow{2}{*}{SimPO} & Original Len. & \heatA{12.5} & \heatB{17.0} & \heatA{11.9} & \heatC{20.0} \\
    & \mname Len. & \heatC{19.9} & \heatF{26.1} & \heatE{24.5} & \heatE{24.2} \\
    \bottomrule
    \end{tabular}
    }
    \caption{Results of the ablation studies. Darker colors indicate higher AlpacaEval 2 LC scores.}
    \label{tab:ablation_detailed}
\end{table}

\subsection{Ablation Studies}
\label{sec:ablation}

To further investigate the impact of the key components of \mname, we conduct experiments on the Mistral-Base (7B) setting \textbf{to demonstrate that simply truncating responses to shorter lengths does not necessarily lead to better performance; instead, the equal-length truncation strategy is crucial for achieving significant gains.}

We compare two truncation strategies:
(1) \textbf{Original Len.}: We truncate both the preferred and dispreferred responses to a certain percentage of their original lengths.
(2) \textbf{\mname Len.}: We first truncate both responses to the length of the shorter one (the \mname strategy), and then further truncate them to a certain percentage of this equal length.
We vary the retention percentage from 25\% to 100\% and report the AlpacaEval 2 results.

The results, summarized in Table~\ref{tab:ablation_detailed}, lead to three main observations.
First, the \textbf{\mname Len.} strategy consistently outperforms the \textbf{Original Len.} strategy across almost all retention ratios for both DPO and SimPO.
Second, the performance gap between the two strategies is already substantial at low retention ratios, indicating that equal-length truncation is more important than simply shortening them.
Third, while simple truncation (Original Len.) brings some gains for DPO, it fails to improve SimPO, whereas the \textbf{\mname Len.} strategy benefits both.
Finally, performance can be further improved by increasing the truncation ratio; nevertheless, as discussed earlier, the parameter-free version of \mname remains the safest choice and already provides strong performance.

\subsection{Comparison with Token-level Methods}

\begin{table}[t]
    \centering
    \begin{tabular}{@{}lccc@{}}
    \toprule
    \textbf{Method} & \textbf{LC} & \textbf{WR} & \textbf{Length} \\
    \midrule
    DPO & 12.9 & 10.6 & 1569 \\
    DPO + SamPO & 22.4 & 15.6 & 1319 \\
    DPO + D$^2$PO & 15.1 & 14.8 & 1925 \\
    \rowcolor{methodbg} DPO + \mname & \best{24.7} & \best{17.7} & 1401 \\
    \midrule
    SimPO & 20.0 & 18.0 & 1777 \\
    SimPO + SamPO & 19.6 & 17.6 & 1789 \\
    SimPO + D$^2$PO & 12.5 & 7.9 & 1127 \\
    \rowcolor{methodbg} SimPO + \mname & \best{24.2} & \best{23.7} & 1939 \\
    \bottomrule
    \end{tabular}
    \caption{Comparison results with other token-level DAAs methods on AlpacaEval 2.}
    \label{tab:token_level_daas}
\end{table}

We compare \mname with two token-level DAAs methods: SamPO~\citep{lu2024eliminating} and D$^2$PO~\citep{shao2025earlier}.
SamPO is designed to mitigate the length bias of DAAs by restricting reward calculation to a random subset of tokens, rather than focusing on the prefix. D$^2$PO prioritizes prefix tokens by applying a temporal decay weight to each token based on its position. Notably, D$^2$PO employs an exponential decay. With a recommended factor of $\gamma=0.98$, the weight for tokens beyond position 200 drops below 0.02, effectively acting as a truncation strategy based on absolute length.

We conduct experiments based on Mistral-7B-Base and report the AlpacaEval 2 results in Table~\ref{tab:token_level_daas}.
As shown in Table~\ref{tab:token_level_daas}, \mname consistently outperforms both SamPO and D$^2$PO across both DPO and SimPO.

\subsection{Safety Alignment Evaluation}
\label{sec:safety_alignment}

We further evaluate the effectiveness of \mname on the safety alignment task.

\paragraph{Setup.}
We utilize the PKU-SafeRLHF dataset~\citep{ji2024pku} for preference optimization, where the safer response is selected as the preferred one. We conduct experiments on two settings: Mistral-Base (7B) and Llama-3-Base (8B).
For evaluation, we use AdvBench~\citep{zou2023universal}, which is a set of 500 harmful behaviors formulated as instructions.
We employ Llama Guard 3-8B~\footnote{\url{https://github.com/meta-llama/PurpleLlama/blob/main/Llama-Guard3/8B/MODEL_CARD.md}} to perform safety classification on the generated responses. The evaluation metric is the safety rate, where a higher rate indicates better performance.

\paragraph{Results.}
The experimental results are presented in Table~\ref{tab:safety_results}.
We observe that \mname achieves substantial improvements over the DPO baseline across settings.
Notably, for the Mistral-Base setting, \mname improves the safety rate from 45.2\% to 82.1\%, a remarkable gain of 36.9 points.
We attribute this significant improvement to the nature of the safety alignment task, where the distinction between safe and unsafe responses typically manifests early in the generation process (e.g., a refusal prefix versus a compliant prefix).
This characteristic aligns perfectly with the design of \mname, which emphasizes the optimization of prefix tokens, thereby effectively bridging the reward-generation gap in safety alignment.

\begin{table}[t]
    \centering
    \resizebox{\columnwidth}{!}{
    \begin{tabular}{lcc}
    \toprule
    \textbf{Method} & \textbf{Mistral-Base (7B)} & \textbf{Llama-3-Base (8B)} \\
    \midrule
    DPO & 45.2 & 78.3 \\
    \rowcolor{methodbg} + \mname & \best{82.1} & \best{87.3} \\
    \bottomrule
    \end{tabular}
    }
    \caption{Safety alignment evaluation results (Safety Rate \%) on PKU-SafeRLHF.}
    \label{tab:safety_results}
\end{table}

\subsection{\mname Generates Better Prefixes}
\label{sec:prefix_quality_analysis}

\begin{figure}[t]
\centering
    \includegraphics[width=0.98\columnwidth]{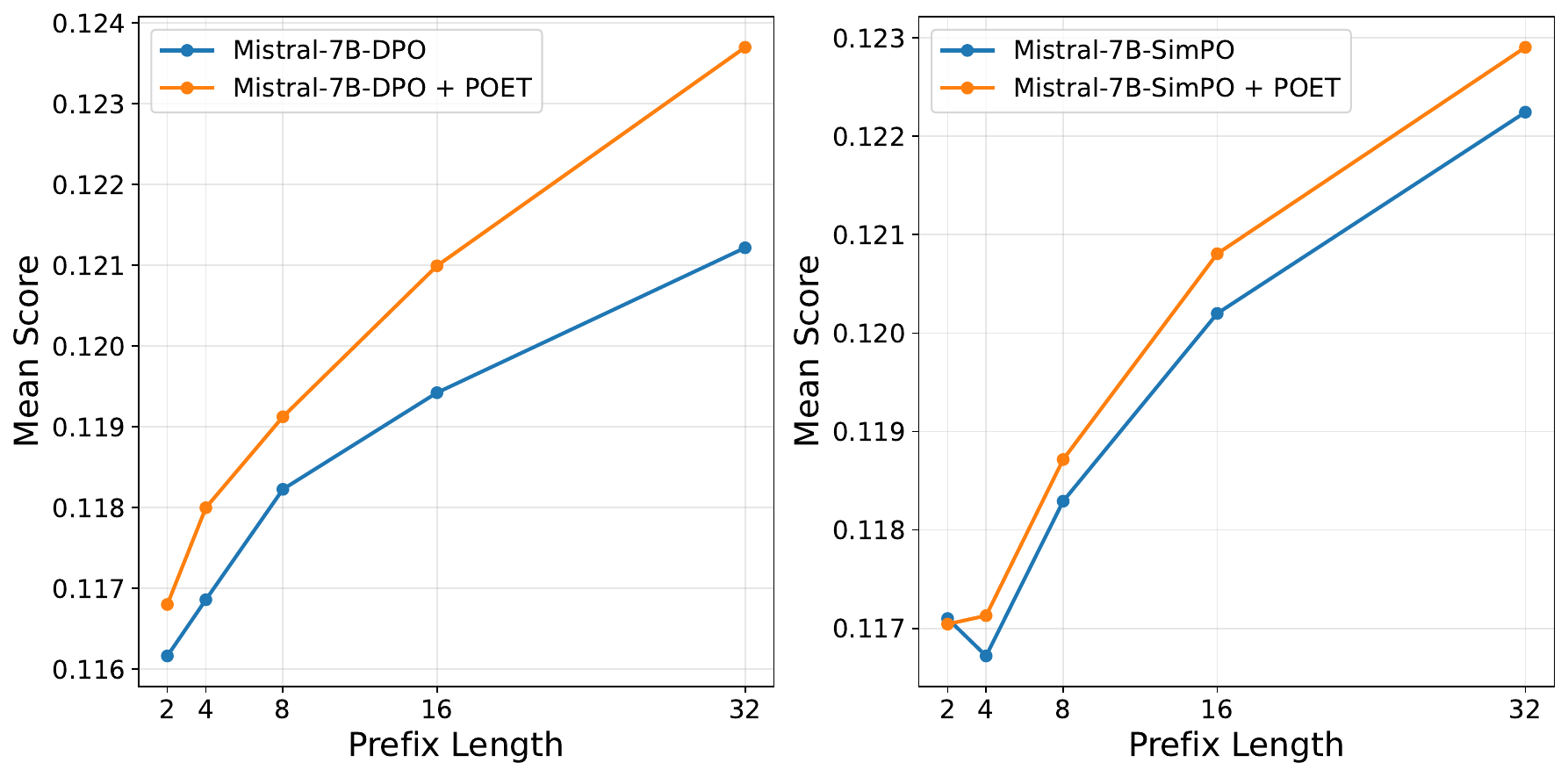}
    \caption{Prefix quality with and without \mname across different prefix lengths. The results show that models trained with \mname consistently generate higher-quality prefixes.}
    \label{fig:prefix_quality}
\end{figure}

To evaluate whether \mname improves performance by generating better prefixes, we conduct an analysis comparing the quality of prefixes generated by standard DAAs models and \mname-trained models. The experimental setup is detailed in Appendix~\ref{app:prefix_quality_analysis}.

\paragraph{Results.}
Figure~\ref{fig:prefix_quality} demonstrates that models trained with \mname consistently generate higher-quality prefixes than standard DAAs models across all prefix lengths, validating the effectiveness of \mname in forcing the DAAs to focus on optimizing the prefix quality, which is the key to bridging the reward-generation gap.

\subsection{When Does \mname Work Best?}
\label{sec:quality_difference_analysis}

\begin{table}[t]
    \centering
    \resizebox{\columnwidth}{!}{
    \begin{tabular}{cccc}
    \toprule
    \textbf{Setting} & \textbf{Consistency (\%)} & \textbf{Difference (*100)} & \textbf{$\Delta$} \\
    \midrule
    \textcircled{1} & 93.8 & 0.6 & -2.8 \\
    \textcircled{2} & 98.5 & 2.3 & +2.0 \\
    \textcircled{3} & 98.9 & 2.9 & +2.9 \\
    \bottomrule
    \end{tabular}
    }
    \caption{Performance comparison on different settings. The Consistency is the quality ranking consistency as in Table~\ref{tab:truncate_acc}. The Difference represents the average reward difference between preferred and dispreferred responses. The $\Delta$ column shows the absolute improvement in AlpacaEval 2 LC.}
    \label{tab:data_analysis}
\end{table}

As discussed in Section~\ref{sec:method}, \mname's effectiveness relies on the practical condition that the quality ranking between preferred and dispreferred responses is preserved after truncation.
In this section, we empirically test this condition and show that the effectiveness of \mname is also connected to the quality difference between preferred and dispreferred responses of the preference dataset.

We conduct experiments using Llama-3-8B and SimPO across three settings with varying degrees of quality difference (details in Appendix~\ref{app:quality_difference_analysis}).
We apply \mname to all three settings and report the results in Table~\ref{tab:data_analysis}. 
The results show that \mname achieves a notable improvement on settings where the consistency rate after applying \mname is high. However, in setting where the consistency rate is only 93.8\%, \mname leads to a performance drop.

Furthermore, we also observe a clear correlation between the quality difference and the effectiveness of \mname.
On one hand, the full response quality difference serves as a good proxy for the quality ranking consistency rate, since the quality difference between preferred and dispreferred responses emerges early in the prefixes (Figure~\ref{fig:prefix_rewards}).
On the other hand, recent researches~\citep{lin2025data,peng2025omni} show that preference samples with larger quality gaps between preferred and dispreferred responses tend to be more informative. Combining our results with those of these works, we conjecture that \mname may perform better on high-quality preference datasets.

\section{Related Work} \label{sec:related_work}

\paragraph{Direct Alignment Algorithms.}
To overcome the instability and computational overhead of RLHF~\citep{zheng2023secrets, santacroce2023efficient}, direct alignment algorithms (DAAs)~\citep{rafailov2024scaling} have been proposed to align LLMs with human preferences without explicit reward modeling or reinforcement learning.
DPO~\citep{Rafailov2023DirectPO} reparameterizes the reward function using the log-ratio between the policy and a reference model, while SimPO~\citep{meng2024simpo} further simplifies this by using length-normalized log-likelihoods, eliminating the need for a reference model.
Other notable DAAs include IPO~\citep{Azar2023AGT}, SLiC-HF~\citep{Zhao2023SLiCHFSL}, ORPO~\citep{Hong2024ORPOMP}, and KTO~\citep{Ethayarajh2024KTOMA}, each proposing different formulations of the preference optimization objective.
Despite their diversity, these methods share a common limitation: their sequence-level objectives may not faithfully reflect autoregressive generation dynamics, which motivates our investigation of the reward-generation gap.

\paragraph{Token-level Preference Optimization.}
\citet{rafailov2024r} showed that DPO can be interpreted through a token-level MDP, establishing a theoretical connection between sequence-level preference optimization and token-level credit assignment.
Building on this insight, several methods have been proposed to explicitly incorporate token-level signals into DAAs.
SamPO~\citep{lu2024eliminating} addresses the length bias of DAAs by computing the reward over a random subset of token positions, effectively decoupling preference learning from response length.
D$^2$PO~\citep{shao2025earlier} applies a temporal decay weight to each token based on its position, assigning exponentially decreasing importance to later tokens in the sequence.
TIS-DPO~\citep{liu2025tisdpo} takes a different approach by estimating token-level importance weights using a separate reward model, performing importance sampling to reweight each token's contribution to the DPO objective.
While these methods modify the training objective to adjust token-level weighting, \mname operates at the data level by truncating response pairs to equal lengths, requiring no changes to the underlying DAA objective and no additional hyperparameters or models.

\section{Conclusion} \label{sec:conclusion}

In this paper, we identify and analyze a critical issue in DAAs—the reward-generation gap, which manifests as a misalignment between optimization objectives during training and autoregressive decoding dynamics.
To address this issue, we introduce \mname, a simple yet effective method for focusing optimization on prefixes by truncating both preferred and dispreferred responses to match the shorter response's length. This hyperparameter-free approach requires no modification to existing DAA objectives while maintaining broad compatibility across different algorithms. Through extensive experiments with DPO and SimPO on multiple model architectures, we demonstrate that our method consistently improves performance.

\section*{Limitations}

The main limitation of \mname is that its effectiveness depends on the preference ordering being preserved after equal-length truncation.
This condition is naturally satisfied when the quality difference between preferred and dispreferred responses emerges early in the generation process.
Consequently, for tasks where the decisive quality signal concentrates at the tail of the sequence (e.g., mathematical reasoning tasks where the final answer token is crucial), the truncation may not preserve the preference ordering, and \mname may not be applicable.
However, it is worth noting that DAAs are currently rarely used for such types of tasks.

\bibliography{custom}

@inproceedings{meng2024simpo,
   title={SimPO: Simple Preference Optimization with a Reference-Free Reward},
   author={Meng, Yu and Xia, Mengzhou and Chen, Danqi},
   booktitle={Advances in Neural Information Processing Systems (NeurIPS)},
   year={2024}
}

@article{rafailov2024scaling,
  title={Scaling laws for reward model overoptimization in direct alignment algorithms},
  author={Rafailov, Rafael and Chittepu, Yaswanth and Park, Ryan and Sikchi, Harshit and Hejna, Joey and Knox, Bradley and Finn, Chelsea and Niekum, Scott},
  journal={arXiv preprint arXiv:2406.02900},
  year={2024}
}

@inproceedings{Rafailov2023DirectPO,
  title     = {Direct Preference Optimization: Your Language Model is Secretly a Reward Model},
  author    = {Rafael Rafailov and Archit Sharma and Eric Mitchell and Stefano Ermon and Christopher D. Manning and Chelsea Finn},
  booktitle = {NeurIPS},
  year      = {2023}
}

@article{Azar2023AGT,
  title   = {A General Theoretical Paradigm to Understand Learning from Human Preferences},
  author  = {Mohammad Gheshlaghi Azar and Mark Rowland and Bilal Piot and Daniel Guo and Daniele Calandriello and Michal Valko and R{\'e}mi Munos},
  journal = {ArXiv},
  year    = {2023},
  volume  = {abs/2310.12036}
}

@article{Ethayarajh2024KTOMA,
  title   = {{KTO}: Model Alignment as Prospect Theoretic Optimization},
  author  = {Kawin Ethayarajh and Winnie Xu and Niklas Muennighoff and Dan Jurafsky and Douwe Kiela},
  journal = {ArXiv},
  year    = {2024},
  volume  = {abs/2402.01306}
}

@article{Hong2024ORPOMP,
  title   = {{ORPO}: Monolithic Preference Optimization without Reference Model},
  author  = {Jiwoo Hong and Noah Lee and James Thorne},
  journal = {ArXiv},
  year    = {2024},
  volume  = {abs/2403.07691}
}

@article{Zhao2023SLiCHFSL,
  title   = {{SLiC-HF}: Sequence Likelihood Calibration with Human Feedback},
  author  = {Yao Zhao and Rishabh Joshi and Tianqi Liu and Misha Khalman and Mohammad Saleh and Peter J. Liu},
  journal = {ArXiv},
  year    = {2023},
  volume  = {abs/2305.10425}
}

@article{Askell2021AGL,
  title   = {A General Language Assistant as a Laboratory for Alignment},
  author  = {Amanda Askell and Yuntao Bai and Anna Chen and Dawn Drain and Deep Ganguli and Tom Henighan and Andy Jones and Nicholas Joseph and Benjamin Mann and Nova DasSarma and Nelson Elhage and Zac Hatfield-Dodds and Danny Hernandez and John Kernion and Kamal Ndousse and Catherine Olsson and Dario Amodei and Tom B. Brown and Jack Clark and Sam McCandlish and Christopher Olah and Jared Kaplan},
  journal = {ArXiv},
  year    = {2021},
  volume  = {abs/2112.00861}
}

@inproceedings{Ouyang2022TrainingLM,
  title     = {Training language models to follow instructions with human feedback},
  author    = {Long Ouyang and Jeff Wu and Xu Jiang and Diogo Almeida and Carroll L. Wainwright and Pamela Mishkin and Chong Zhang and Sandhini Agarwal and Katarina Slama and Alex Ray and John Schulman and Jacob Hilton and Fraser Kelton and Luke E. Miller and Maddie Simens and Amanda Askell and Peter Welinder and Paul Francis Christiano and Jan Leike and Ryan J. Lowe},
  booktitle = {NeurIPS},
  year      = {2022}
}

@article{Tunstall2023ZephyrDD,
  title   = {Zephyr: Direct Distillation of {LM} Alignment},
  author  = {Lewis Tunstall and Edward Beeching and Nathan Lambert and Nazneen Rajani and Kashif Rasul and Younes Belkada and Shengyi Huang and Leandro von Werra and Cl{\'e}mentine Fourrier and Nathan Habib and Nathan Sarrazin and Omar Sanseviero and Alexander M. Rush and Thomas Wolf},
  journal = {ArXiv},
  year    = {2023},
  volume  = {abs/2310.16944}
}

@inproceedings{zheng2023judging,
  title     = {Judging {LLM}-as-a-Judge with {MT-Bench} and {Chatbot} {Arena}},
  author    = {Zheng, Lianmin and Chiang, Wei-Lin and Sheng, Ying and Zhuang, Siyuan and Wu, Zhanghao and Zhuang, Yonghao and Lin, Zi and Li, Zhuohan and Li, Dacheng and Xing, Eric and others},
  booktitle = {NeurIPS Datasets and Benchmarks Track},
  year      = {2023}
}

@misc{arenahard2024,
  title  = {From Live Data to High-Quality Benchmarks: The {Arena-Hard} Pipeline},
  url    = {https://lmsys.org/blog/2024-04-19-arena-hard/},
  author = {Tianle Li and Wei-Lin Chiang and Evan Frick and Lisa Dunlap and Banghua Zhu and Joseph E. Gonzalez and Ion Stoica},
  month  = {April},
  year   = {2024}
}

@misc{AlpacaEval,
  author       = {Xuechen Li and Tianyi Zhang and Yann Dubois and Rohan Taori and Ishaan Gulrajani and Carlos Guestrin and Percy Liang and Tatsunori B. Hashimoto },
  title        = {{AlpacaEval}: An Automatic Evaluator of Instruction-following Models},
  year         = {2023},
  publisher    = {GitHub},
  journal      = {GitHub repository},
  howpublished = {\url{https://github.com/tatsu-lab/alpaca_eval}}
}

@article{dubois2024length,
  title   = {Length-Controlled {AlpacaEval}: A Simple Way to Debias Automatic Evaluators},
  author  = {Dubois, Yann and Galambosi, Bal{\'a}zs and Liang, Percy and Hashimoto, Tatsunori B},
  journal = {ArXiv},
  volume  = {abs/2404.04475},
  year    = {2024}
}

@article{llama3modelcard,
  title  = {Llama 3 Model Card},
  author = {AI@Meta},
  year   = {2024},
  url    = {https://github.com/meta-llama/llama3/blob/main/MODEL_CARD.md}
}

@inproceedings{Ding2023EnhancingCL,
  title     = {Enhancing Chat Language Models by Scaling High-quality Instructional Conversations},
  author    = {Ning Ding and Yulin Chen and Bokai Xu and Yujia Qin and Zhi Zheng and Shengding Hu and Zhiyuan Liu and Maosong Sun and Bowen Zhou},
  booktitle = {EMNLP},
  year      = {2023}
}

@inproceedings{Cui2024UltraFeedbackBL,
  title     = {{UltraFeedback}: Boosting Language Models with High-quality Feedback},
  author    = {Ganqu Cui and Lifan Yuan and Ning Ding and Guanming Yao and Wei Zhu and Yuan Ni and Guotong Xie and Zhiyuan Liu and Maosong Sun},
  year      = {2024},
  booktitle = {ICML}
}

@article{ziegler2019fine,
  title   = {Fine-tuning language models from human preferences},
  author  = {Ziegler, Daniel M and Stiennon, Nisan and Wu, Jeffrey and Brown, Tom B and Radford, Alec and Amodei, Dario and Christiano, Paul and Irving, Geoffrey},
  journal = {arXiv preprint arXiv:1909.08593},
  year    = {2019}
}

@article{zheng2023secrets,
  title   = {Secrets of {RLHF} in Large Language Models Part {I: PPO}},
  author  = {Zheng, Rui and Dou, Shihan and Gao, Songyang and Hua, Yuan and Shen, Wei and Wang, Binghai and Liu, Yan and Jin, Senjie and Liu, Qin and Zhou, Yuhao and others},
  journal = {arXiv preprint arXiv:2307.04964},
  year    = {2023}
}

@article{singhal2023long,
  title   = {A Long Way to Go: Investigating Length Correlations in {RLHF}},
  author  = {Singhal, Prasann and Goyal, Tanya and Xu, Jiacheng and Durrett, Greg},
  journal = {arXiv preprint arXiv:2310.03716},
  year    = {2023}
}

@article{santacroce2023efficient,
  title   = {Efficient {RLHF}: Reducing the Memory Usage of {PPO}},
  author  = {Santacroce, Michael and Lu, Yadong and Yu, Han and Li, Yuanzhi and Shen, Yelong},
  journal = {arXiv preprint arXiv:2309.00754},
  year    = {2023}
}

@article{bai2022training,
  title   = {Training a helpful and harmless assistant with reinforcement learning from human feedback},
  author  = {Bai, Yuntao and Jones, Andy and Ndousse, Kamal and Askell, Amanda and Chen, Anna and DasSarma, Nova and Drain, Dawn and Fort, Stanislav and Ganguli, Deep and Henighan, Tom and others},
  journal = {arXiv preprint arXiv:2204.05862},
  year    = {2022}
}

@article{christiano2017deep,
  title   = {Deep reinforcement learning from human preferences},
  author  = {Christiano, Paul F and Leike, Jan and Brown, Tom and Martic, Miljan and Legg, Shane and Amodei, Dario},
  journal = {Advances in neural information processing systems},
  volume  = {30},
  year    = {2017}
}

@inproceedings{Brown2020LanguageMA,
  title     = {Language Models are Few-Shot Learners},
  author    = {Brown, Tom and Mann, Benjamin and Ryder, Nick and Subbiah, Melanie and Kaplan, Jared D and Dhariwal, Prafulla and Neelakantan, Arvind and Shyam, Pranav and Sastry, Girish and Askell, Amanda and others},
  booktitle = {NeurIPS},
  year      = {2020}
}

@article{stiennon2020learning,
  title   = {Learning to summarize with human feedback},
  author  = {Stiennon, Nisan and Ouyang, Long and Wu, Jeffrey and Ziegler, Daniel and Lowe, Ryan and Voss, Chelsea and Radford, Alec and Amodei, Dario and Christiano, Paul F},
  journal = {Advances in Neural Information Processing Systems},
  volume  = {33},
  pages   = {3008--3021},
  year    = {2020}
}

@inproceedings{li-etal-2016-deep,
  title     = {Deep Reinforcement Learning for Dialogue Generation},
  author    = {Li, Jiwei  and
               Monroe, Will  and
               Ritter, Alan  and
               Jurafsky, Dan  and
               Galley, Michel  and
               Gao, Jianfeng},
  editor    = {Su, Jian  and
               Duh, Kevin  and
               Carreras, Xavier},
  booktitle = {Proceedings of the 2016 Conference on Empirical Methods in Natural Language Processing},
  month     = nov,
  year      = {2016},
  address   = {Austin, Texas},
  publisher = {Association for Computational Linguistics},
  url       = {https://aclanthology.org/D16-1127},
  doi       = {10.18653/v1/D16-1127},
  pages     = {1192--1202}
}

@article{cobbe2021gsm8k,
  title   = {Training Verifiers to Solve Math Word Problems},
  author  = {Cobbe, Karl and Kosaraju, Vineet and Bavarian, Mohammad and Chen, Mark and Jun, Heewoo and Kaiser, Lukasz and Plappert, Matthias and Tworek, Jerry and Hilton, Jacob and Nakano, Reiichiro and Hesse, Christopher and Schulman, John},
  journal = {arXiv preprint arXiv:2110.14168},
  year    = {2021}
}

@misc{open-llm-leaderboard,
  author       = {Edward Beeching and Clémentine Fourrier and Nathan Habib and Sheon Han and Nathan Lambert and Nazneen Rajani and Omar Sanseviero and Lewis Tunstall and Thomas Wolf},
  title        = {Open {LLM} Leaderboard},
  year         = {2023},
  publisher    = {Hugging Face},
  howpublished = {\url{https://huggingface.co/spaces/HuggingFaceH4/open_llm_leaderboard}}
}

@article{xu2024contrastive,
  title   = {Contrastive Preference Optimization: Pushing the Boundaries of {LLM} Performance in Machine Translation},
  author  = {Haoran Xu and Amr Sharaf and Yunmo Chen and Weiting Tan and Lingfeng Shen and Benjamin Van Durme and Kenton Murray and Young Jin Kim},
  journal = {ArXiv},
  year    = {2024},
  volume  = {abs/2401.08417}
}

@article{pal2024smaug,
  title   = {Smaug: Fixing Failure Modes of Preference Optimisation with {DPO}-Positive},
  author  = {Pal, Arka and Karkhanis, Deep and Dooley, Samuel and Roberts, Manley and Naidu, Siddartha and White, Colin},
  journal = {arXiv preprint arXiv:2402.13228},
  year    = {2024}
}

@inproceedings{ArmoRM,
  title   = {Interpretable Preferences via Multi-Objective Reward Modeling and Mixture-of-Experts},
  author  = {Haoxiang Wang and Wei Xiong and Tengyang Xie and Han Zhao and Tong Zhang},
  booktitle = {Findings of EMNLP},
  year    = {2024}
}

@article{team2024gemma,
  title   = {Gemma 2: Improving open language models at a practical size},
  author  = {Team, Gemma and Riviere, Morgane and Pathak, Shreya and Sessa, Pier Giuseppe and Hardin, Cassidy and Bhupatiraju, Surya and Hussenot, L{\'e}onard and Mesnard, Thomas and Shahriari, Bobak and Ram{\'e}, Alexandre and others},
  journal = {arXiv preprint arXiv:2408.00118},
  year    = {2024}
}

@article{shi2024understanding,
  title={Understanding likelihood over-optimisation in direct alignment algorithms},
  author={Shi, Zhengyan and Land, Sander and Locatelli, Acyr and Geist, Matthieu and Bartolo, Max},
  journal={arXiv preprint arXiv:2410.11677},
  year={2024}
}

@article{ranzato2015sequence,
  title={Sequence level training with recurrent neural networks},
  author={Ranzato, Marc'Aurelio and Chopra, Sumit and Auli, Michael and Zaremba, Wojciech},
  journal={arXiv preprint arXiv:1511.06732},
  year={2015}
}

@inproceedings{arora-etal-2022-exposure,
    title = "Why Exposure Bias Matters: An Imitation Learning Perspective of Error Accumulation in Language Generation",
    author = "Arora, Kushal  and
      El Asri, Layla  and
      Bahuleyan, Hareesh  and
      Cheung, Jackie",
    editor = "Muresan, Smaranda  and
      Nakov, Preslav  and
      Villavicencio, Aline",
    booktitle = "Findings of the Association for Computational Linguistics: ACL 2022",
    month = may,
    year = "2022",
    address = "Dublin, Ireland",
    publisher = "Association for Computational Linguistics",
    url = "https://aclanthology.org/2022.findings-acl.58/",
    doi = "10.18653/v1/2022.findings-acl.58",
    pages = "700--710"
}

@inproceedings{zhang-etal-2021-trading,
    title = "Trading Off Diversity and Quality in Natural Language Generation",
    author = "Zhang, Hugh  and
      Duckworth, Daniel  and
      Ippolito, Daphne  and
      Neelakantan, Arvind",
    editor = "Belz, Anya  and
      Agarwal, Shubham  and
      Graham, Yvette  and
      Reiter, Ehud  and
      Shimorina, Anastasia",
    booktitle = "Proceedings of the Workshop on Human Evaluation of NLP Systems (HumEval)",
    month = apr,
    year = "2021",
    address = "Online",
    publisher = "Association for Computational Linguistics",
    url = "https://aclanthology.org/2021.humeval-1.3/",
    pages = "25--33"
}

@misc{jiang2023mistral7b,
      title={Mistral 7B}, 
      author={Albert Q. Jiang and Alexandre Sablayrolles and Arthur Mensch and Chris Bamford and Devendra Singh Chaplot and Diego de las Casas and Florian Bressand and Gianna Lengyel and Guillaume Lample and Lucile Saulnier and Lélio Renard Lavaud and Marie-Anne Lachaux and Pierre Stock and Teven Le Scao and Thibaut Lavril and Thomas Wang and Timothée Lacroix and William El Sayed},
      year={2023},
      eprint={2310.06825},
      archivePrefix={arXiv},
      primaryClass={cs.CL},
      url={https://arxiv.org/abs/2310.06825}, 
}

@article{shao2024deepseekmath,
  title={Deepseekmath: Pushing the limits of mathematical reasoning in open language models},
  author={Shao, Zhihong and Wang, Peiyi and Zhu, Qihao and Xu, Runxin and Song, Junxiao and Bi, Xiao and Zhang, Haowei and Zhang, Mingchuan and Li, YK and Wu, Y and others},
  journal={arXiv preprint arXiv:2402.03300},
  year={2024}
}

@article{zhou2023instruction,
  title={Instruction-following evaluation for large language models},
  author={Zhou, Jeffrey and Lu, Tianjian and Mishra, Swaroop and Brahma, Siddhartha and Basu, Sujoy and Luan, Yi and Zhou, Denny and Hou, Le},
  journal={arXiv preprint arXiv:2311.07911},
  year={2023}
}

@article{chen2021evaluating,
  title={Evaluating large language models trained on code},
  author={Chen, Mark and Tworek, Jerry and Jun, Heewoo and Yuan, Qiming and Pinto, Henrique Ponde de Oliveira and Kaplan, Jared and Edwards, Harri and Burda, Yuri and Joseph, Nicholas and Brockman, Greg and others},
  journal={arXiv preprint arXiv:2107.03374},
  year={2021}
}

@article{roziere2023code,
  title={Code llama: Open foundation models for code},
  author={Roziere, Baptiste and Gehring, Jonas and Gloeckle, Fabian and Sootla, Sten and Gat, Itai and Tan, Xiaoqing Ellen and Adi, Yossi and Liu, Jingyu and Sauvestre, Romain and Remez, Tal and others},
  journal={arXiv preprint arXiv:2308.12950},
  year={2023}
}

@article{touvron2023llama,
  title={Llama: Open and efficient foundation language models},
  author={Touvron, Hugo and Lavril, Thibaut and Izacard, Gautier and Martinet, Xavier and Lachaux, Marie-Anne and Lacroix, Timoth{\'e}e and Rozi{\`e}re, Baptiste and Goyal, Naman and Hambro, Eric and Azhar, Faisal and others},
  journal={arXiv preprint arXiv:2302.13971},
  year={2023}
}

@article{ji2025first,
  title={The first few tokens are all you need: An efficient and effective unsupervised prefix fine-tuning method for reasoning models},
  author={Ji, Ke and Xu, Jiahao and Liang, Tian and Liu, Qiuzhi and He, Zhiwei and Chen, Xingyu and Liu, Xiaoyuan and Wang, Zhijie and Chen, Junying and Wang, Benyou and others},
  journal={arXiv preprint arXiv:2503.02875},
  year={2025}
}

@article{dubois2023alpacafarm,
  title={Alpacafarm: A simulation framework for methods that learn from human feedback},
  author={Dubois, Yann and Li, Chen Xuechen and Taori, Rohan and Zhang, Tianyi and Gulrajani, Ishaan and Ba, Jimmy and Guestrin, Carlos and Liang, Percy S and Hashimoto, Tatsunori B},
  journal={Advances in Neural Information Processing Systems},
  volume={36},
  pages={30039--30069},
  year={2023}
}

@inproceedings{lin2025data,
  title={Data with high and consistent preference difference are better for reward model},
  author={Lin, Qi and Lu, Hengtong and Yuan, Caixia and Wang, Xiaojie and Jiang, Huixing and Chen, Wei},
  booktitle={Proceedings of the AAAI Conference on Artificial Intelligence},
  volume={39},
  number={26},
  pages={27482--27490},
  year={2025}
}

@article{peng2025omni,
  title={Omni-DPO: A Dual-Perspective Paradigm for Dynamic Preference Learning of LLMs},
  author={Peng, Shangpin and Wang, Weinong and Tian, Zhuotao and Yang, Senqiao and Wu, Xing and Xu, Haotian and Zhang, Chengquan and Isobe, Takashi and Hu, Baotian and Zhang, Min},
  journal={arXiv preprint arXiv:2506.10054},
  year={2025}
}

@article{rafailov2024r,
  title={From $r$ to $q^{*}$: Your language model is secretly a q-function},
  author={Rafailov, Rafael and Hejna, Joey and Park, Ryan and Finn, Chelsea},
  journal={arXiv preprint arXiv:2404.12358},
  year={2024}
}

@article{hendryckstest2021,
  title={Measuring Massive Multitask Language Understanding},
  author={Dan Hendrycks and Collin Burns and Steven Basart and Andy Zou and Mantas Mazeika and Dawn Song and Jacob Steinhardt},
  journal={Proceedings of the International Conference on Learning Representations (ICLR)},
  year={2021}
}

@inproceedings{lu2024eliminating,
  title={Eliminating Biased Length Reliance of Direct Preference Optimization via Down-Sampled KL Divergence},
  author={Lu, Junru and Li, Jiazheng and An, Siyu and Zhao, Meng and He, Yulan and Yin, Di and Sun, Xing},
  booktitle={Proceedings of the 2024 Conference on Empirical Methods in Natural Language Processing},
  pages={1047--1067},
  year={2024}
}

@inproceedings{shao2025earlier,
  title={Earlier Tokens Contribute More: Learning Direct Preference Optimization From Temporal Decay Perspective},
  author={Ruichen Shao and Bei Li and Gangao Liu and Yang Chen and ZhouXiang and Jingang Wang and Xunliang Cai and Peng Li},
  booktitle={The Thirteenth International Conference on Learning Representations},
  year={2025},
  url={https://openreview.net/forum?id=OspqtLVUN5}
}

@inproceedings{liu2025tisdpo,
  title={{TIS}-{DPO}: Token-level Importance Sampling for Direct Preference Optimization With Estimated Weights},
  author={Aiwei Liu and Haoping Bai and Zhiyun Lu and Yanchao Sun and Xiang Kong and Xiaoming Simon Wang and Jiulong Shan and Albin Madappally Jose and Xiaojiang Liu and Lijie Wen and Philip S. Yu and Meng Cao},
  booktitle={The Thirteenth International Conference on Learning Representations},
  year={2025},
  url={https://openreview.net/forum?id=oF6e2WwxX0}
}

@article{ji2024pku,
  title={PKU-SafeRLHF: Towards Multi-Level Safety Alignment for LLMs with Human Preference},
  author={Ji, Jiaming and Hong, Donghai and Zhang, Borong and Chen, Boyuan and Dai, Josef and Zheng, Boren and Qiu, Tianyi and Li, Boxun and Yang, Yaodong},
  journal={arXiv preprint arXiv:2406.15513},
  year={2024}
}

@article{zou2023universal,
  title={Universal and transferable adversarial attacks on aligned language models},
  author={Zou, Andy and Wang, Zifan and Carlini, Nicholas and Nasr, Milad and Kolter, J Zico and Fredrikson, Matt},
  journal={arXiv preprint arXiv:2307.15043},
  year={2023}
}

\appendix

\section{Dynamics of Token-level Entropies and Log Probabilities}
\label{app:token_dynamics}
\begin{figure*}[htbp]
\centering
    \subfigure[]{
            \includegraphics[width=0.48\textwidth]{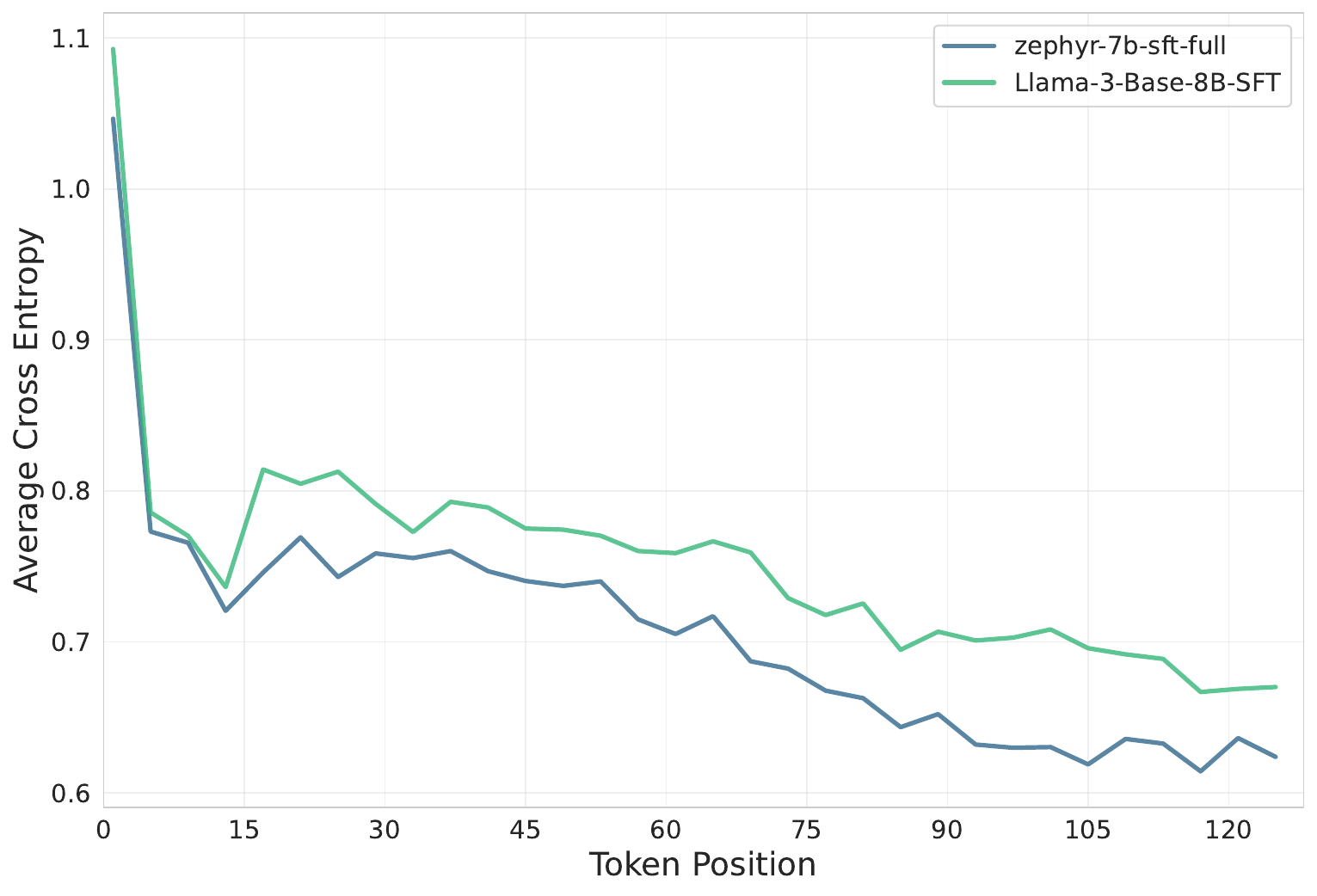}
            \label{fig:token_entropy}
    }
    \hfill
    \subfigure[]{
        \includegraphics[width=0.48\textwidth]{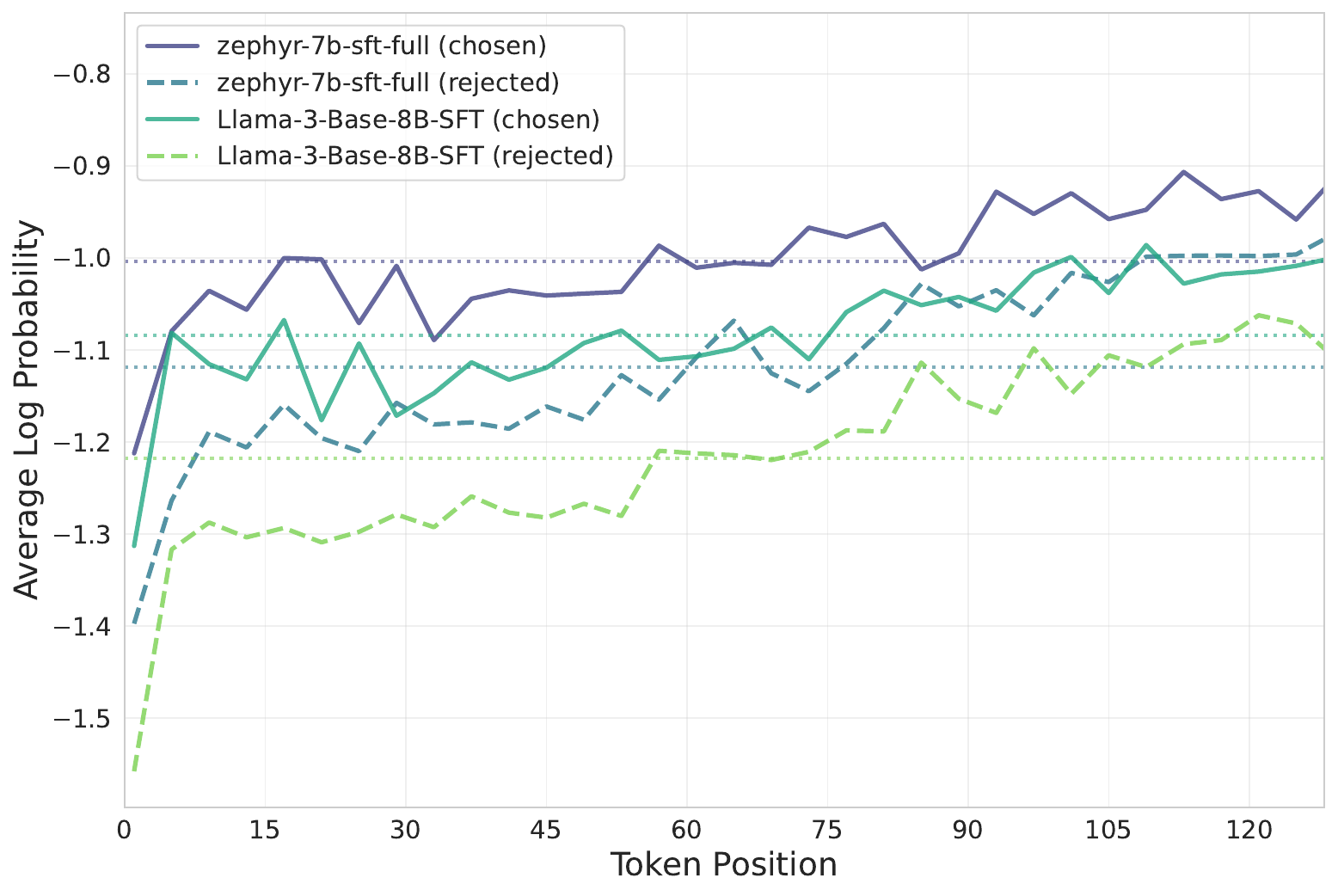}
        \label{fig:token_lopg}
    }
    \caption{(a) Average per-position cross entropy when sampling from prompts. (b) Average per-position log probability. The horizontal line represents the mean log probability across all positions.}
\label{fig:logp_and_entropy}
\end{figure*}

In this section, we present additional empirical evidence on the dynamics of token-level entropies and log probabilities in Figure~\ref{fig:logp_and_entropy}, to support our analysis in Section~\ref{sec:issue}.
Figure~\ref{fig:token_entropy} illustrates the average per-position cross entropy when sampling from prompts of the UltraFeedback test set.
We observe that prefix tokens have significantly higher uncertainty compared to later tokens, as the latter benefit from more context and exhibit lower randomness.
Furthermore, Figure~\ref{fig:token_lopg} presents the average per-position log probability across all responses in the UltraFeedback test set.
It reveals that prefix tokens exhibit significantly lower log probabilities.
Despite this, DAAs' implicit reward functions treat all tokens equally, which dilutes the importance of these critical early tokens due to the overwhelming number of subsequent tokens.

\section{Proof of Theorem~\ref{thm:optimality}}
\label{appendix:theorem_proof}

\begin{proof}
Following the insights from~\citet{rafailov2024r}, for $v \in \{w, l\}$, define:
\begin{equation}
R_v(k) = \sum_{t=0}^{k} \beta \log \frac{\pi^*(\mathbf{a}_t^v \mid \mathbf{s}_t^v)}{\pi_{\mathrm{ref}}(\mathbf{a}_t^v \mid \mathbf{s}_t^v)},
\end{equation}
where $\pi^*$ is the optimal policy under the MDP. From Lemma 1 in~\citet{rafailov2024r}, we obtain the token-level reward decomposition for a partial sequence up to step $k$:

\begin{equation}
\sum_{t=0}^{k} r\left(\mathbf{s}_t^v, \mathbf{a}_t^v\right) = V^*(\mathbf{s}_0^v) + R_v(k) - V^*(\mathbf{s}_{k+1}^v).
\end{equation}

Substituting into Eq.~\ref{eq:el_bt_model}, we have:

\begin{align}
    &\sum_{t=0}^{k} r\left(\mathbf{s}_t^v, \mathbf{a}_t^v\right) + V^*(\mathbf{s}_{k+1}^v) \nonumber\\
    &= V^*(\mathbf{s}_0^v) + R_v(k) - V^*(\mathbf{s}_{k+1}^v) + V^*(\mathbf{s}_{k+1}^v) \nonumber\\
    &= V^*(\mathbf{s}_0^v) + R_v(k).
\end{align}

Since both trajectories start from the same prompt $x$, we have $V^*(\mathbf{s}_0^w) = V^*(\mathbf{s}_0^l)$. Therefore, the equal-length sub-trajectories BT model simplifies to:
\begin{equation}
    p^*_k(y_{w,\leq k} \succeq y_{l,\leq k}) = \sigma\big(R_w(k) - R_l(k)\big),
\end{equation}
where $\sigma$ is the sigmoid function. This is equivalent to the DPO loss function expressed through the optimal policy, demonstrating that the equal-length sub-trajectory BT model and the original sequence-level BT model yield the same optimal policy.

\end{proof}

\section{Downstream Experimental Results}
\label{app:downstream_tasks}

Table~\ref{tab:all_downstream} lists detailed results of each downstream task.

\begin{table*}[t]
    \resizebox{\textwidth}{!}{
    \begin{tabular}{@{}lccccccc@{}}
    \toprule
    & \textbf{MMLU (5)} & \textbf{ARC (25)} & \textbf{HellaSwag (10)} & \textbf{TruthfulQA (0)} & \textbf{Winograd (5)} & \textbf{GSM8K (5)} & \textbf{Average} \\
    \midrule
    \multicolumn{8}{c}{\textbf{Mistral-Base (7B)}} \\ 
    \midrule
    SFT   & 58.93          & 57.59         & 80.65               & 40.35                & 76.72             & 34.34          & 58.10        \\ 
    \midrule
    DPO   & 58.77          & 62.37         & 83.91               & 47.76                & 76.72             & 29.34          & 59.81        \\ 
    + \mname & 57.93          & 64.25         & 84.48               & 54.68                & 76.95             & 30.33          & 61.44        \\ 
    \midrule
    SimPO & 57.71          & 62.29         & 83.43               & 51.08                & 77.74             & 30.17          & 60.40        \\ 
    + \mname & 58.12        & 62.54         & 83.39               & 50.85                & 77.74             & 34.50          & 61.19        \\ \midrule
    \multicolumn{8}{c}{\textbf{Llama-3-Base (8B)}} \\
    \midrule
    SFT   & 63.79          & 60.15         & 81.59               & 45.32                & 76.32             & 50.80          & 62.99        \\
    \midrule
    DPO   & 63.50          & 63.74         & 83.86               & 53.67                & 76.80             & 53.60          & 65.86        \\
    + \mname & 63.45          & 65.53         & 83.98               & 55.25                & 76.40             & 51.10          & 65.95        \\
    \midrule
    SimPO & 63.15          & 65.02         & 82.90               & 59.78                & 77.66             & 48.98          & 66.25        \\
    + \mname & 63.34        & 64.93         & 82.82               & 58.20                & 77.66             & 54.21          & 66.86        \\ 
    \midrule
    \multicolumn{8}{c}{\textbf{Llama-3-Instruct v0.2 (8B)}} \\
    \midrule
    SFT   & 61.3 & 78.8 & 51.6 & 65.7 & 76.5 & 75.9 & 68.3 \\
    \midrule
    DPO   & 64.8 & 79.9 & 56.2 & 65.9 & 76.6 & 77.4 & 70.1 \\
    + \mname & 64.1 & 79.2 & 56.6 & 65.9 & 76.3 & 75.4 & 69.6 \\
    \midrule
    SimPO & 67.2 & 78.5 & 65.6 & 65.1 & 76.4 & 68.5 & 70.2 \\
    + \mname & 66.5 & 79.0 & 63.4 & 65.6 & 77.1 & 69.9 & 70.2 \\ 
    \midrule
    \multicolumn{8}{c}{{\textbf{Gemma-2-Instruct (9B)}}} \\
    \midrule
    SFT   & 71.1 & 81.8 & 60.1 & 72.3 & 78.5 & 48.9 & 68.8 \\
    \midrule
    DPO   & 69.5 & 71.6 & 57.9 & 72.5 & 72.1 & 41.9 & 64.3 \\
    + \mname & 70.6 & 71.6 & 57.5 & 72.4 & 72.7 & 45.0 & 65.0 \\
    \midrule
    SimPO & 69.1 & 67.3 & 59.1 & 71.9 & 73.5 & 40.6 & 63.6 \\
    + \mname & 69.2 & 68.2 & 60.5 & 71.6 & 74.0 & 44.1 & 64.6 \\
    \bottomrule
    \end{tabular}
    }
    \caption{Downstream task evaluation results of tasks on the huggingface open leaderboard.}
    \label{tab:all_downstream}
\end{table*}

\section{Hyperparameters}
\label{app:hyperparameters}

Table~\ref{tab:hyperparameters} summarizes the hyperparameters used for the four main experimental settings reported in the main body of the paper.
We follow the hyperparameter configurations of DPO and SimPO as described in~\citep{meng2024simpo}. In all experiments, we set max\_prompt\_length to 1800 and max\_length to 2048.

\begin{table*}[htbp]
    \centering
    \resizebox{\textwidth}{!}{
    \begin{tabular}{llcccc}
    \toprule
    \textbf{Method} & \textbf{Parameter} & \textbf{Mistral-Base (7B)} & \textbf{Llama-3-Base (8B)} & \textbf{Llama-3-Instruct v0.2 (8B)} & \textbf{Gemma-2-Instruct (9B)} \\
    \midrule
    \multirow{2}{*}{DPO} & $\beta$ & 0.01 & 0.01 & 0.01 & 0.01 \\
    & Learning rate & $5\text{e-}7$ & $5\text{e-}7$ & 3e-7 & 5e-7 \\
    \midrule
    \multirow{3}{*}{SimPO} & $\beta$ & 2.0 & 2.0 & 10 & 10 \\
    & $\gamma / \beta$ & 0.8 & 0.5 & 0.3 & 0.5 \\
    & Learning rate & $3\text{e-}7$ & $6\text{e-}7$ & $1\text{e-}6$ & $8\text{e-}7$ \\
    \bottomrule
    \end{tabular}
    }
    \caption{Hyperparameters used in our main experiments.}
    \label{tab:hyperparameters}
\end{table*}

\section{Evaluations}
\label{app:evals}

The official implementation of AlpacaEval 2~\citep{dubois2024length} uses the \textit{weighted\_alpaca\_eval\_gpt4\_turbo} annotator, which employs the \textit{gpt-4-1106-preview} as backbone model. In our experiments, we maintain the same annotator methodology (weighted win rate) but substituted the backbone model with \textit{deepseek-v3-0324}, which is substantially cheaper than \textit{gpt-4-1106-preview} while achieving better performance. We analyze our annotator using AlpacaEval 2's \textit{analyze\_evaluators} command and compare it with the official annotator in Table~\ref{tab:annotator_comparison}. As shown, our annotator achieves higher human agreement, Spearman correlation, and Pearson correlation, while being significantly cheaper.
Since Arena-Hard-v0.1~\citep{arenahard2024} also employs \textit{gpt-4-1106-preview} as its backbone evaluation model, we substitute it with \textit{deepseek-v3-0324} for the same reasons. 

For benchmarks from the Huggingface Open Leaderboard~\citep{open-llm-leaderboard}, we use \textit{lm-evaluation-harness} for evaluation.

\section{Implementation Details} \label{app:implementation_details}
All the training experiments in this paper were conducted on 8$\times$A800 GPUs based on the alignment-handbook repo and the codebase of \citep{meng2024simpo}.

\section{Experimental Setup for Prefix Quality Analysis}
\label{app:prefix_quality_analysis}

\begin{table*}[ht]
    \centering
    \resizebox{\textwidth}{!}{
    \begin{tabular}{@{}lcccccc@{}}
    \toprule
    \textbf{Annotator} & \textbf{Human Agreement} & \textbf{Price} & \textbf{Spearman Corr.} & \textbf{Pearson Corr.} & \textbf{Bias} & \textbf{Variance} \\
    \midrule
    weighted\_alpaca\_eval\_deepseek\_v3\_0324 & 67.27 & 0.12 & 0.95 & 0.87 & 32.19 & 16.45 \\
    weighted\_alpaca\_eval\_gpt4\_turbo & 65.73 & 4.32 & 0.78 & 0.77 & 33.90 & 23.65 \\
    \bottomrule
    \end{tabular}
    }
    \caption{Comparison of AlpacaEval 2 Annotators.}
    \label{tab:annotator_comparison}
\end{table*}

In this section, we provide a detailed experimental setup for the prefix quality analysis in Section~\ref{sec:prefix_quality_analysis}.

We randomly sample 500 prompts from the test set of UltraFeedback and generate 5 responses per prompt using both standard DAAs and \mname-trained models. These responses are then truncated to create prefixes of varying lengths. Following the definition of prefix quality in Eq.~\ref{eq:prefix_quality}, we evaluate the quality of these prefixes by generating completions from them using the SFT model as proxy policy. This choice of proxy policy isolates the influence of other factors.
For each prefix, we use the SFT model to generate 3 completions. As a result, we obtain 15 completions (5 responses with 3 completions each) for each prompt at each prefix length. We then evaluate the quality of these completions using a strong reward model, ArmoRM-Llama3-8B-v0.1~\citep{ArmoRM}.

\section{Experimental Setup for Quality Difference Analysis}
\label{app:quality_difference_analysis}
In this section, we provide a detailed experimental setup for the analysis in Section~\ref{sec:quality_difference_analysis}.

We conduct an analysis using Llama-3-8B and SimPO across 3 settings with varying degrees of quality difference between preferred and dispreferred responses:

\begin{enumerate}[leftmargin=*]
\item[\textcircled{1}] \textbf{Llama-3-8B-SimPO v0.1}~\citep{meng2024simpo}: In this setting, the responses in preference dataset are generated by Llama-3-8B-Instruct~\citep{llama3modelcard} using the prompts from the UltraFeedback dataset, with preference annotating using a weak reward model. This setting is expected to have a small quality difference between preferred and dispreferred responses of the preference dataset.
\item[\textcircled{2}] \textbf{Llama-3-8B-SimPO v0.2}~\citep{meng2024simpo}: Same as the v0.1 setting, but with a strong reward model, RLHFlow/ArmoRM-Llama3-8B-v0.1~\citep{ArmoRM}, used for preference annotating. This setting is expected to have a moderate quality difference.
\item[\textcircled{3}] \textbf{Llama-3-8B-SimPO Interpolation}: In this setting, we create two new models by interpolating between Llama-3-8B and Llama-3-8B-Instruct with ratios of 0.1:0.9 and -0.1:1.1, respectively. We then generate responses by the two interpolated models and the Llama-3-8B-Instruct model (2 responses per model), following the pipeline with the Llama-3-8B-SimPO-v0.2 setting. This setting is expected to have a large quality difference, as the generated responses are more diverse than the previous two settings.
\end{enumerate}

The quality of preferred and dispreferred responses is estimated by ArmoRM-Llama3-8B-v0.1~\citep{ArmoRM}.

\end{document}